\documentclass[sigconf,screen]{acmart}
\AtBeginDocument{%
  \providecommand\BibTeX{{%
    \normalfont B\kern-0.5em{\scshape i\kern-0.25em b}\kern-0.8em\TeX}}}

\setcopyright{acmlicensed}
\copyrightyear{2019}
\acmYear{2019}
\acmConference[RecSys '19]{Thirteenth ACM Conference on Recommender Systems}{September 16--20, 2019}{Copenhagen, Denmark}
\acmBooktitle{Thirteenth ACM Conference on Recommender Systems (RecSys '19), September 16--20, 2019, Copenhagen, Denmark}
\acmPrice{15.00}
\acmDOI{10.1145/3298689.3347055}
\acmISBN{978-1-4503-6243-6/19/09}

\newcommand{\F}{\mathbb{R}}
\newcommand{\Fb}{\mathbb{B}}

\newcommand{\eref}[1]{Eq.~(\ref{#1})}
\newcommand{\sref}[1]{Sec.~\ref{#1}}

\newcommand{\tref}[1]{Table~\ref{#1}}

\newcommand{\Sref}[1]{Section~\ref{#1}}

\renewcommand\vec{\mathbf}
\DeclareMathOperator{\diag}{diag}

\begin{document}

\title{HybridSVD: When Collaborative Information is Not Enough}

\author{Evgeny Frolov}
\orcid{0000-0003-3679-5311}
\affiliation{%
  \institution{Skolkovo Institute of Science and Technology}
  \city{Moscow}
  \country{Russia}
}
\email{evgeny.frolov@skoltech.ru}

\author{Ivan Oseledets}
\additionalaffiliation{%
  \institution{Institute of Numerical Mathematics, Russian Academy of Sciences}
  \city{Moscow}
  \country{Russia}
}
\affiliation{%
  \institution{Skolkovo Institute of Science and Technology}
  \city{Moscow}
  \country{Russia}
}
\email{i.oseledets@skoltech.ru}

\renewcommand{\shortauthors}{E. Frolov and I. Oseledets}

\begin{abstract}
We propose a new hybrid algorithm that allows incorporating both user and item side information within the standard collaborative filtering technique. One of its key features is that it naturally extends a simple PureSVD approach and inherits its unique advantages, such as highly efficient Lanczos-based optimization procedure, simplified hyper-parameter tuning and a quick folding-in computation for generating recommendations instantly even in highly dynamic online environments. The algorithm utilizes a generalized formulation of the singular value decomposition, which adds flexibility to the solution and allows imposing the desired structure on its latent space. Conveniently, the resulting model also admits an efficient and straightforward solution for the cold start scenario. We evaluate our approach on a diverse set of datasets and show its superiority over similar classes of hybrid models.

\end{abstract}



\maketitle

\section{Introduction}\label{intro}
Singular value decomposition (SVD) \cite{golub2012matrix} is a well-established and useful tool with a wide range of applications in various domains of information retrieval, natural language processing, and data analysis. The first SVD-based collaborative filtering (CF) models were proposed in the late 90's early 00's and were successfully adapted to a wide range of tasks \cite{billsus1998learning,sarwar2000application,kim2005collaborative}. It has also anticipated active research devoted to alternative matrix factorization (MF) techniques \cite{koren2009matrix}. However, despite the development of many sophisticated and accurate methods, the simplest SVD-based approach called \emph{PureSVD} was later shown to outperform other algorithms in standard top-$n$ recommendation tasks \cite{Cremonesi2010}.

PureSVD offers a unique set of practical advantages, such as global convergence with deterministic output, lightweight hyper-parameter tuning via simple rank truncation, an analytical expression for instant online recommendations (see Section~\ref{subsec:efficiency}), scalable modifications based on randomized algorithms \cite{halko2011finding}. It has highly optimized implementations in many programming languages based on BLAS and LAPACK routines and is included in many modern machine learning libraries and frameworks, some of which allow to handle nearly billion-scale problems\footnote{https://github.com/criteo/Spark-RSVD}.

However, like any other conventional collaborative filtering technique, PureSVD relies solely on the knowledge about user preferences expressed in the form of ratings, likes, purchases or other types of feedback, either explicit or implicit. On the other hand, a user's choice may be influenced by intrinsic properties of items. For example, users may prefer products of a particular category/brand or products with certain characteristics. Similarly, additional knowledge about users, such as demographic information or occupation, may also help to explain their choice.

These influencing factors are typically assumed to be well represented by the model's latent features.
However, in situations when users interact with a small number of items from a large assortment (e.g., in online stores), it may become difficult to build reliable models from the observed behavior without considering side information. This additional knowledge may also help in the extreme \emph{cold start} \cite{ekstrand2011collaborative} case, when preference data for some items and/or users are not available at all.

Addressing the described problems of insufficient preference data has lead to the development of hybrid models \cite{burke2002hybrid}.
A plethora of ways for building hybrid recommenders has been explored to date, and a vast amount of work is devoted specifically to incorporating side information into MF methods (more on this in Section~\ref{sec:related}). Surprisingly, despite many of its advantages, the \emph{PureSVD model has received much lower attention} in this regard.
To the best of our knowledge, there were no attempts for developing an integrated hybrid approach, where \emph{interactions data and side information would be jointly factorized with the help of SVD}, and the obtained result would be used as an \emph{end model} for generating recommendations immediately as in the PureSVD case.

With this work, we aim to fill that gap and extend the family of hybrid algorithms with a new approach based on a modification of PureSVD. The main contributions of this paper are:

\begin{itemize}
    \item We generalize PureSVD by allowing it to incorporate and to jointly factorize side information along with collaborative data. We call this approach \emph{HybridSVD}.
    \item We propose an efficient computational scheme that does not break the simplicity of the algorithm, requires minimum efforts in parameter tuning, and leads to a minor increase in overall complexity governed by a structure of input data.
    \item We introduce a simple yet surprisingly capable approach for cold start scenario, which can be implemented for both HybridSVD and PureSVD models.
\end{itemize}

The rest of this paper is organized as follows. We walk through SVD internals in \Sref{sec:svd_limits} and show what exactly leads to the limitations of its standard implementation. In \Sref{sec:model} we demonstrate how to remove these limitations by switching to an algebraic formulation of the generalized SVD.
\Sref{sec:experiments} provides evaluation methodology with model selection process and the results are reported in \Sref{sec:results}. \Sref{sec:related} describes the related research and \Sref{sec:conclusion} concludes the work.

\section{Understanding SVD limitations}\label{sec:svd_limits}
One of the greatest advantages of many MF methods is their flexibility in terms of a problem formulation. It allows specifying very fine-grained models that incorporate both interactions data and additional knowledge within a single optimization objective (some of these methods are described in \sref{sec:related}). It often results in the formation of a latent feature space with a particular inner structure, controlled by side information.

This technique, however, is not available off-the-shelf in the PureSVD approach due to the classical formulation with its fixed least squares objective. In this work, we aim to find a new way to formulate the optimization problem so that, while staying within the computational paradigm of SVD, it would allow us to account for additional sources of information during the optimization process.

\subsection{When PureSVD does not work}\label{subsec:example}
Consider the following simple example on fictitious interactions data depicted in \tref{tab:cold_start}. Initially, we have 3 users (Alice, Bob and Carol) and 5 items, with only the first 4 items being observed in interactions (the first 3 rows and 4 columns of the table). Item in the last column (\emph{Item5}) plays the role of a cold-start (i.e., previously unobserved) item. We use this toy data to build PureSVD of rank 2 and generate recommendations for a new user Tom (\emph{New user} row in the table), who has already interacted with \emph{Item1}, \emph{Item4}, \emph{Item5}.

Suppose that in addition to interaction data we know that \emph{Item3 and Item5 are more similar to each other} than to other items in terms of some set of features. In that case, since Tom has expressed an interest in \emph{Item5}, it seems natural to expect from a sound recommender system to favor \emph{Item3} over \emph{Item2} in recommendations. This, however, does not happen with PureSVD.

With the help of the \emph{folding-in} technique \cite{ekstrand2011collaborative} (also see \eref{eq:folding-in-std})
it can be easily verified, that SVD will assign uniform scores for both items as shown in the \emph{PureSVD} row in the bottom of the table. This example illustrates a general limitation of the PureSVD approach related to the lack of preference data, which can not be resolved without considering side information (or unless more data is collected). In contrast, our approach will assign a higher score to \emph{Item3} (see \emph{Our approach} row in \tref{tab:cold_start} as an example), which reflects the relations between \emph{Item3} and \emph{Item5}.

\begin{table}[t]
    \caption{An example of insufficient preference data problem}
    \label{tab:cold_start}
    \begin{minipage}{\columnwidth}
        \begin{center}
            \begin{tabular}{l c c c c c}
                \multicolumn{1}{c}{} & Item1 & Item2 & \color{blue} Item3 & Item4 & \color{blue} Item5 \\ \cline{2-6}
                \multicolumn{1}{c}{} & \multicolumn{5}{c}{\textit{Observed interactions}} \\ \hline
                Alice & 1 &   & 1 & 1 &  \\
                Bob   & 1 & 1 &   & 1 &  \\
                Carol & 1 &   &   & 1 &  \\
                \multicolumn{1}{c}{} & \multicolumn{5}{c}{\textit{New user}} \\ \hline
                Tom   & 1 & ? & ? & 1 & 1 \\
                \multicolumn{1}{c}{} & \multicolumn{4}{c}{\textit{}} & \multicolumn{1}{c}{}\\ 

                \multicolumn{2}{r}{PureSVD:} & \multicolumn{1}{c}{0.3} & \multicolumn{1}{c}{0.3} & \multicolumn{1}{c}{} & \multicolumn{1}{c}{} \\

                \multicolumn{2}{r}{Our approach:} & \multicolumn{1}{c}{0.1} & \multicolumn{1}{c}{0.6} & \multicolumn{1}{c}{} & \multicolumn{1}{c}{} \\
            \end{tabular}
        \end{center}
        \bigskip
        \small{\emph{Table note:} \emph{Item3} and \emph{Item5} are {\color{blue} similar} to each other (as indicated by {\color{blue} blue} color). \emph{PureSVD} fails to take that into account and assigns uniform scores to Item3 and Item2 for Tom.
        \emph{Our approach} favors Item3 as it is similar to previous Tom's choice. The code to reproduce this result can be found at \href{https://gist.github.com/Evfro/c6ec2b954adfff6aaa356f9b3124b1d2}{https://gist.github.com/Evfro/c6ec2b954adfff6aaa356f9b3124b1d2}.}
    \end{minipage}
\end{table}

\subsection{Why PureSVD does not work}
Formal explanation of the observed result requires an understanding of how exactly computations are performed in the SVD algorithm. A rigorous mathematical analysis of that is performed by the authors of the EIGENREC model \cite{eigenrec}. As the authors note, the \emph{latent factor model of PureSVD can be viewed as an eigendecomposition of a scaled user-based or item-based cosine similarity matrix}. We, therefore, would expect it to depend on scalar products between rows and columns of a user-item interactions matrix.

Indeed, in the user-based case, it solves an eigendecomposition problem for the Gram matrix
\begin{equation}\label{eq:gram}
    \vec{G} = \vec{RR}^\top,
\end{equation}
where $\vec{R} \in \F^{M \times N}$ is a matrix of interactions between $M$ users and $N$ items with unknown elements imputed by zeros (as required by PureSVD). The elements of $\vec{G}$ are defined by scalar products of the corresponding rows of $\vec{R}$ that represent user preferences:
\begin{equation}\label{eq:svd_cosine}
    g_{ij} = \vec{r}_i^\top\vec{r}_j.
\end{equation}
This observation immediately suggests that \emph{any cross-item relations are simply ignored by PureSVD} as it takes only item co-occurrence into account. That is, \emph{the contribution of a particular item into the user similarity score $g_{ij}$ is counted only if the item is present in the preferences of both user $i$ and user $j$}. The similar conclusion also holds for the item-based case. This explains the uniform scores assigned by PureSVD in our fictitious example in \tref{tab:cold_start}.

\section{Proposed model}\label{sec:model}
In order to account for cross-entity relations, we have to find a different similarity measure that would consider all possible pairs of entities and allow injecting side information. It can be achieved by replacing the scalar product in \eref{eq:svd_cosine} with a bilinear form:
\begin{equation}\label{eq:outer}
    g_{ij}  = \vec{r}_i^\top \vec{S} \, \vec{r}_j.
\end{equation}
where symmetric matrix $\vec{S} \in \F^{N \times N}$ reflects auxiliary relations between items based on side information.
Effectively, this matrix creates ``virtual'' links between users even if they have no common items in their preferences.
Occasional links will be filtered out by dimensionality reduction, whereas more frequent ones will help to reveal valuable consumption patterns.

In a similar fashion we can introduce matrix $\vec{K} \in \F^{M \times M}$ to incorporate user-related information. We will use the term \emph{side similarity} to denote these matrices. Their off-diagonal entries encode similarities between users or items based on side information, such as user attributes or item features. We discuss the properties of such matrices and possible ways of constructing them in \sref{subsec:sidesim}.

\emph{Our primary goal is to bring them together into a joint problem formulation with a single solution based on SVD}.

\subsection{HybridSVD}
Replacing scalar products with bilinear forms as per \eref{eq:outer} for all users and all items generates the following matrix cross-products:
\begin{equation}\label{eq:crossproduct}
    \vec{RSR}^\top, \quad \vec{R}^\top \vec{KR},
\end{equation}
where $\vec{R}$ is the same is in PureSVD. By analogy with standard SVD, this induces the following eigendecomposition problem:
\begin{equation}
    \label{eq:problem}
    \begin{cases}
        \vec{RSR}^\top = \vec{U\Sigma}^2\vec{U}^\top,\\
        \vec{R}^\top\vec{KR} = \vec{V\Sigma}^2\vec{V}^\top,
    \end{cases}
\end{equation}
where matrices $\vec{U} \in \F^{M \times k}$ and $\vec{V} \in \F^{N \times k}$ represent embeddings of users and items onto a \emph{common $k$-dimensional latent feature space}. This system of equations has a close connection to the \emph{Generalized SVD} \cite{golub2012matrix}
and can be solved via the standard SVD of an auxiliary matrix $\vec{\widehat R}$ \cite{gsvd} of the form:
\begin{equation}\label{eq:gsvd}
    \vec{\widehat R} \equiv \vec{K}^{\frac{1}{2}}\vec{R\,S}^{\frac{1}{2}} = \vec{\widehat U}\vec{\Sigma}\vec{\widehat V}^\top,
\end{equation}
where matrices $\vec{\widehat U}$, $\vec{\Sigma}$ and $\vec{\widehat V}$ represent singular triplets of $\vec{\widehat R}$ similarly to the PureSVD model. Connections between the original and the auxiliary latent space for both users and items are established by $\vec{U} = \vec{K}^{-1/2}\vec{\widehat U}$ and $\vec{V} = \vec{S}^{-1/2}\vec{\widehat V}$ respectively.

We note, however, that finding the square root of an arbitrary matrix and its inverse are computationally intensive operations. To overcome that difficulty we will assume by now that matrices $\vec{K}$ and $\vec{S}$ are symmetric positive definite (more details on that in \sref{subsec:sidesim}) and therefore can be represented in the \emph{Cholesky decomposition} form: $\vec{S} = \vec{L}_S^{}\vec{L}_S^\top$, $\vec{K} = \vec{L}_K^{}\vec{L}_K^\top$, where $\vec{L}_S$ and $\vec{L}_K$ are lower triangular real matrices. This decomposition can be computed much more efficiently than finding matrix square root.

By a direct substitution, it can be verified that the following auxiliary matrix
\begin{equation}
    \label{eq:cholesky}
    \vec{\widehat R} \equiv \vec{L}_K^\top\,\vec{R\,L}_S^{} = \vec{\widehat U\Sigma}\vec{\widehat V}^\top
\end{equation}
also provides a solution to \eref{eq:problem} and can be used to replace \eref{eq:gsvd} with its expensive square root computation. The connection between the auxiliary latent space and the original one now becomes
\begin{equation}\label{eq:latentspace}
    \vec{U} = \vec{L}_K^{-\top} \vec{\widehat U}, \quad
    \vec{V} = \vec{L}_S^{-\top} \vec{\widehat V}.
\end{equation}
As can be seen, matrix $\vec{\widehat R}$ ``absorbs'' additional relations encoded by matrices $\vec{K}$ and $\vec{S}$ simultaneously. It shows that \emph{solving the joint problem in \eref{eq:problem} is as simple as finding standard SVD of an auxiliary matrix from \eref{eq:cholesky}}.
We call this model \emph{HybridSVD}.

The orthogonality of singular vectors $\vec{\widehat U}^\top\vec{\widehat U} = \vec{\widehat V}^\top\vec{\widehat V} = \vec{I}$ combined with \eref{eq:latentspace} also sheds the light on the effect of side information on the structure of hybrid latent space, revealing its central property:
\begin{equation}\label{eq:orthogon}
    \vec{U}^\top \vec{KU} = \vec{V}^\top \vec{SV} = \vec{I},
\end{equation}
i.e., columns of the matrices $\vec{U}$ and $\vec{V}$ are orthogonal under the constraints imposed by the matrices $\vec{K}$ and $\vec{S}$ respectively\footnote{ This property is called  $\vec{K}$- and $\vec{S}$-orthogonality \cite{strangbook}.}.

\subsection{Side similarity}\label{subsec:sidesim}
In order to improve the quality of collaborative models, side similarity matrices must capture representative structure from side information. Constructing such matrices is a domain-specific task, and there are numerous ways for achieving this, ranging from a direct application of various similarity measures over the feature space \cite{lu2011link} to more specialized techniques based on kernel methods, Graph Laplacians, matrix factorization, and deep learning \cite{cai2018comprehensive}.

Independently of the way similarity matrices are obtained, we impose the following structure on them:
\begin{equation}\label{eq:similarity}
   \begin{cases}
       \vec{S} = (1-\alpha)\,\vec{I} + \alpha\,\vec{Z}, \\
       \vec{K} = (1-\beta)\,\vec{I} + \beta\,\vec{Y},
   \end{cases}
\end{equation}
where $\vec{Z},\vec{Y}$ are real symmetric matrices of conforming size with entries encoding actual feature-based similarities and taking values from $\left[-1, 1\right]$ interval.
Coefficients $\alpha, \beta \in \left[0, 1\right]$ are free model parameters determined empirically. Adjusting these coefficients \emph{changes the balance between collaborative and side information}: higher values put more emphasis on side features, whereas lower values gravitate the model towards simple co-occurrence patterns.

Note that in the case of $\alpha=\beta=0$ the model turns back into PureSVD. On the other extreme, when $\alpha=\beta=1$ the model would heavily rely on feature-based similarities, which can be an undesired behavior. Indeed, if an underlying structure of side features is less expressive than that of collaborative information (which is not a rare case), the model is likely to suffer from overspecialization.
Hence, the values of $\alpha$ and $\beta$ will be most often lower than 1.
Moreover, setting them below a certain threshold\footnote{An upper bound can be estimated from the matrix \emph{diagonal dominance} condition \cite{golub2012matrix}.} would ensure positive definiteness of $\vec{S}$ and $\vec{K}$, giving a unique Cholesky decomposition.

In this work, we explore one of the most direct possible approaches for constructing similarity matrices. We transform categorical side features (e.g., movie genre or product brand, see \tref{tab:datasets}) into one-hot encoded vectors and compute the Common Neighbors similarity \cite{lu2011link} scaled by a normalization constant. More specifically, in the item case, we form a sparse binary matrix $\vec{F} \in \Fb^{N \times f}$, which rows are one-hot feature vectors: if a feature belongs to an item, then the corresponding entry in the row is 1, otherwise 0.
The feature-based similarity matrix is then computed as $\vec{Z} = \frac{1}{\left|m\right|}\vec{FF}^\top$, where $m$ denotes the largest matrix element of $\vec{FF}^\top$. We additionally enforce the diagonal elements of $\vec{Z}$ to be all ones: $\vec{Z} \leftarrow \vec{Z} + \vec{I} - \diag\diag\vec{Z}$.

\subsection{Efficient computations}\label{subsec:efficiency}
\paragraph{Cholesky Decomposition}
Sparse feature representation allows having sparse matrices $\vec{K}$ and $\vec{S}$. This can be exploited for computing \emph{sparse Cholesky decomposition} or, even better, incomplete Cholesky decomposition \cite{golub2012matrix}, additionally allowing to skip negligibly small similarity values. The corresponding triangular part of the Cholesky factors will be sparse as well. Moreover, \emph{there is no need to explicitly calculate inverses} of $\vec{L}_S^\top$ and $\vec{L}_K^\top$ in \eref{eq:latentspace} as it only requires to solve the corresponding triangular system of equations, which can be performed very efficiently \cite{golub2012matrix}.

Furthermore, Cholesky decomposition is fully deterministic and admits symbolic factorization. This feature makes tuning HybridSVD even more efficient. Indeed, changing the values of $\alpha$ and $\beta$ in $(0, 1)$ interval does not affect the sparsity structure of \eref{eq:similarity}. Therefore, once $\vec{L}_S$ and $\vec{L}_K$ are computed for some values of $\alpha$ and $\beta$, the resulting symbolic sparsity pattern can be reused later to perform quick calculations with other values of these coefficients. We used the CHOLMOD library \cite{chen2008cholmod} for that purpose, available in the scikit-sparse package\footnote{https://github.com/scikit-sparse/scikit-sparse}.
Computing \emph{sparse} Cholesky decomposition was orders of magnitude faster than computing the model itself.

In general, however, \emph{side features do not have to be categorical or even sparse}. For example, if item features are represented by compact embeddings encoded as rows of a dense matrix $\vec{F}\in\F^{N \times f}$ with $f\!\ll\!N$, one can utilize the ``identity plus low rank`` structure of similarity matrices and apply fast symmetric factorization techniques \cite{ambikasaran2014symfact} to efficiently compute matrix roots. The inverses can then be computed via Sherman-Woodbury-Morrison formula. We leave the investigation of this alternative for future research.

\paragraph{Computing SVD} There is also no need to directly compute the product of Cholesky factors and rating matrix in \eref{eq:cholesky}, which could potentially lead to a dense result. Similarly to the standard SVD, assuming the rank of decomposition $k \ll \min{\left(M, N\right)}$, one can exploit the Lanczos procedure \cite{lanczos1950iteration,golub2012matrix}, which invokes a Krylov subspace method. The key benefit of such approach is that in order to find $k$ principal singular triplets it is sufficient to provide a rule of multiplying an auxiliary matrix from \eref{eq:cholesky} by an arbitrary dense vector from the left and the right. This can be implemented as a sequence of 3 matrix-vector multiplications. Hence, the added computational cost of the algorithm over standard SVD is controlled by the complexity of multiplying Cholesky factors by a dense vector.

More specifically, given the number of non-zero elements $nnz_R$ of $\vec{R}$ that corresponds to the number of the observed interactions, an overall computational complexity of the HybridSVD algorithm is \mbox{$O(nnz_R \cdot k) + O((M+N) \cdot k^2) + O((J_K + J_S)\cdot k)$}, where the first 2 terms correspond to PureSVD's complexity and the last term depends on the complexities $J_K$ and $J_S$ of multiplying matrices $\vec{L}_K$ and $\vec{L}_S$ by a dense vector.
In our case $J_K$ and $J_S$ are determined by the number of non-zero elements $nnz_{L_K}$ and $nnz_{L_S}$ of the sparse Cholesky factors. Therefore, the total complexity amounts to \mbox{$O(nnz_{tot}\!\cdot\!k)\!+\!O((M+N)\!\cdot\!k^2)$}, where \mbox{$nnz_{tot}\!=\!nnz_R\!+\!nnz_{L_K}\!+\!nnz_{L_S}$}.

\paragraph{Generating recommendations} One of the greatest advantages of the SVD-based approach is analytical form of the \emph{folding-in} \cite{ekstrand2011collaborative}. In contrast to many other MF methods, it is stable and does not require any additional optimization steps to calculate recommendations in the warm start regime. It makes SVD-based models especially plausible for highly dynamic online environments.
For example, in the case of a new ``warm'' user with some vector of known preferences $\vec{p}$ a vector of predicted item relevance scores $\vec{\tilde r}$ reads:
\begin{equation}\label{eq:folding-in-std}
    \vec{\tilde r} \approx \vec{\widetilde V}\vec{\widetilde V}^\top \, \vec{p},
\end{equation}
where $\vec{\widetilde V}$ corresponds to the right singular vectors of the PureSVD model. Moreover, it turns into strict equality in the case of a \emph{known user} \cite{Cremonesi2010}. Therefore, it presents a \emph{single solution for generating recommendations for both known and new users}.

This result can also be generalized to the hybrid case.
Following the same folding-in idea and taking into account the special structure of the latent space given by \eref{eq:latentspace} one arrives at the following expression for the vector $\vec{r}$ of predicted item scores:
\begin{equation}\label{eq:folding-in}
    \vec{r} \approx \vec{L}_S^{-\top}\vec{\widehat V}\vec{\widehat V}^\top \vec{L}_S^\top\vec{p} = \vec{V}_l^{} \vec{V}_r^\top \vec{p},
\end{equation}
where $\vec{V}_l^{} = \vec{L}_S^{-\top}\vec{\widehat V}$ and $\vec{V}_r^{} = \vec{L}_S\vec{\widehat V}$. As with PureSVD, it is suitable for predicting preferences of both known and warm-start users.

There is no $\vec{K}$ matrix in \eref{eq:folding-in} due to the nature of the folding-in approximation. If needed, one can put more emphasis on user features by combining the folding-in vector with the cold start representation (see \sref{subsec:coldstart}). We, however, do not investigate this option and leave it for future work. In our setup, described in \sref{sec:experiments}, only item features are available. Hence, no modification is required.

\subsection{Cold start settings}\label{subsec:coldstart}
Unlike the warm start, in the cold start settings, information about preferences is not known at all. Handling this situation depends on the ability to translate known user attributes and item characteristics into the latent feature space. \emph{We will focus on the item cold start scenario}. User cold start can be managed in a similar fashion. The task is to find among known users the ones who would be the most interested in a new cold item. Hence, in order to utilize a standard prediction scheme via scalar products of latent vectors, one needs to obtain latent features of the cold item.

We propose to solve the task in two steps. We start by finding a linear map $\vec{W}$ between real features of known items and their latent representations. We define the corresponding problem as
\begin{equation}\label{eq:coldmap}
        \vec{VW} = \vec{F},
\end{equation}
where rows of $\vec{F}$ encode item features. This step is performed right after the model is computed, and the result is stored for later computations.
Secondly, once the mapping $\vec{W}$ is obtained, we use it to transform any cold item represented by its feature vector $\vec{f}$ into the corresponding latent feature vector $\vec{v}$ by solving a linear system:
\begin{equation}
        \vec{W}^\top\vec{v} = \vec{f}.
\end{equation}
Finally, the prediction vector can be computed as
\begin{equation}\label{eq:coldstart}
    \vec{r} = \vec{U \Sigma v} = \vec{RV v}.
\end{equation}

The issue with \eref{eq:coldmap} is that solving it for $\vec{W}$ can be a challenging task. Hence, many hybrid methods incorporate feature mapping into the main optimization objective, which allows solving two problems simultaneously.
However, the orthogonality property of HybridSVD defined by \eref{eq:orthogon} admits a direct solution to \eref{eq:coldmap}:
\begin{equation}
    \vec{W} = \vec{V}^\top \vec{SF}.
\end{equation}
This result also provides solution for PureSVD by setting $\vec{S}=\vec{I}$. We use this technique in our cold start experiments to verify whether HybridSVD provides any benefit over the PureSVD approach.

\subsection{Matrix scaling}\label{subsec:scaling}
According to the EIGENREC model \cite{eigenrec}, a simple scaling trick
\begin{equation}\label{eq:scaling}
    \vec{R} \leftarrow \vec{R} \vec{D}^{d-1},
\end{equation}
where $\vec{D}\!=\!\diag\{\|\vec{\bar{r}}_1\|, \ldots, \|\vec{\bar{r}}_N\|\}$ contains Euclidean norms of the columns $\vec{\bar{r}}_i$ of $\vec{R}$, \emph{can significantly improve the quality of PureSVD}. In the default setting the scaling factor $d$ is equal to 1, giving the standard model. Varying its value governs popularity biases: higher values would emphasize the significance of popular items, whereas lower values would increase sensitivity to rare ones.
Empirically, the values slightly below 1 performed best. In our experiments, we employ this technique for both HybridSVD and PureSVD and report results in addition to the original non-scaled models.

\section{Experiments}\label{sec:experiments}
We conduct two sets of experiments: \emph{standard top-$n$ recommendation scenario} and \emph{item cold start scenario}.
Every experiment starts with a hyperparameter tuning phase with $n=10$ fixed. Once the optimal parameter values are found, they are used for the final evaluation of recommendations quality for a range of values of $n$. Experiments were performed using the \emph{Polara framework}\footnote{https://github.com/Evfro/polara}. The source code for reproducing all results is available in our public Github repository\footnote{https://github.com/Evfro/recsys19\_hybridsvd}.

\subsection{Evaluation methodology}
In the \emph{standard scenario} we consequently mark every 20\% of users for test. Each 20\% partition contains only those users who have not been tested yet. We randomly withdraw a single item from every test user and collect all such items in a holdout set. After that, the test users are merged back with the remaining 80\% of users and form a training set. During the evaluation phase, we generate a ranked list of top-$n$ recommendations for every test user based on their known preferences and evaluate it against the holdout.

In the \emph{cold start scenario} we perform 80\%/20\% partitioning of the list of all unique items. We select items from a 20\% partition and mark them as cold-start items. Users with at least one cold-start item in the preferences are marked as the test users. Users with no items in their preferences, other than cold-start items, are filtered out. The remaining users form a training set with all cold-start items excluded. Evaluation of models, in that case, is performed as follows: for every cold-start item, we generate a ranked list of the most pertinent users and evaluate it against one of the test users chosen randomly among those who have interacted with the item.

In both \emph{standard} and \emph{cold start} experiments we conventionally perform 5-fold cross-validation and average the results, providing 95\% confidence intervals based on the paired t-test criteria.
The quality of recommendations is measured with the help of \emph{mean reciprocal rank} (MRR) \cite{kluver2018rating} metric. We also report \emph{hit-rate} (HR) \cite{deshpande2004item} and \emph{coverage} -- a fraction of all unique recommended entities to the total amount of unique entities in the training set. The latter characterizes an overall diversity of recommendations.

\subsection{Datasets}\label{subsec:datasets}
We have used 5 different publicly available datasets: \emph{MovieLens-1M} (ML1M) and \emph{MovieLens-10M} (ML10M) \cite{harper2016movielens}; \emph{BookCrossing} (BX) \cite{ziegler2005bookcross}; \emph{Amazon Electonics} (AMZe) and \emph{Amazon Video Games} (AMZvg) \cite{mcauley2015amazon}; large scale \emph{R2 Yahoo! Music} (YaMus) \cite{yahoomusic}.

In all datasets, we ensure that for every item at least 5 users have interacted with it, and every user has interacted with at least 5 items. We also remove items without any side information.
As we are not interested in the rating prediction, the setting with \emph{only binary feedback} is considered in our experiments. We use only categorical item features for side similarity computation, following the technique described at the end of \sref{subsec:sidesim}. The main dataset characteristics after data preprocessing are provided in \tref{tab:datasets}.

Accordingly, both Movielens and Amazon datasets are binarized with a threshold value of 4, i.e., lower ratings are removed, and the remaining ratings are set to 1.
In the case of Movielens datasets, we have extended \emph{genres} data with \emph{cast, directors and writers} information crawled from the TMDB database\footnote{https://www.themoviedb.org}.
For Amazon datasets, we used information about \emph{category hierarchy} and \emph{brands}. In order to avoid too dense similarity matrices, we use only the lowest level hierarchies. In the case of AMZvg dataset, we additionally remove the ``Games'' category due to its redundancy.

In the BX dataset, we select only the part with implicit feedback.
We additionally filter out users with more than 1000 rated items. Information about \emph{authors} and \emph{publishers} provided within the dataset is used to build side similarity matrices.
In the case of Yahoo!Music dataset, due to its size, we take only 10\% of data corresponding to the first CV split, provided by Yahoo Labs. We select only interactions with rating 5 and binarize them. Side similarity is computed using information about \emph{genres, artists} and \emph{albums}.

\begin{table}
\caption{Description of the datasets after pre-processing.}
\centering{
\begin{tabular}{l*{4}{c}}
\hline
  & \textbf{\# users} & \textbf{\# items} & \textbf{nnz} & \textbf{side information}\\
\hline
\textbf{ML-1M} & 6038 & 3522 & 2.7\% & genres, cast, \\
\textbf{ML-10M} & 69797 & 10258 & 0.7\% & directors, writers \\
\hline
\textbf{BX} & 7160 & 16273 & 0.18\% & publishers, authors \\
\hline
\textbf{AMZe} & 124895 & 44483 & 0.02\% & categories, \\
\textbf{AMZvg} & 14251 & 6858 & 0.13\% & brands \\
\hline
\textbf{YaMus} & 183003 & 134059 & 0.1\% & artists, genres, albums \\
\hline
\end{tabular}
}
\label{tab:datasets}
\end{table}

\begin{figure*}[ht]
\centering
\includegraphics[width=\textwidth]{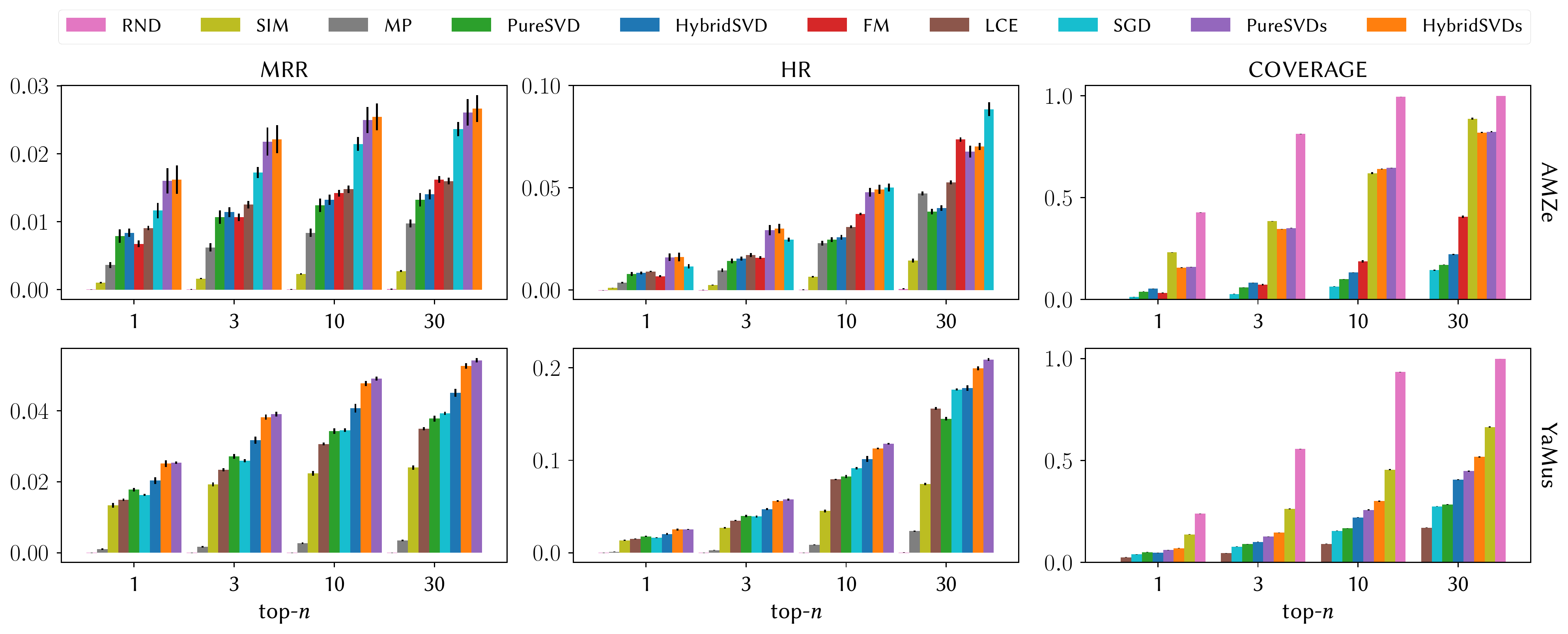}
\caption{Standard scenario for the Amazon Electronics (top row) and Yahoo! Music (bottom row) datasets. Generally, the scaled versions of the SVD-based algorithms outperform the competing methods. Side information is less important than scaling.}
\label{fig:std_mrr}
\end{figure*}

\subsection{Baseline algorithms}\label{subsec:baselines}
We compare our approach against several standard baseline models as well as against two hybrid models that generalize many existing techniques. Below is a brief description of each competing method.

\emph{PureSVD} is a simple SVD-based model that replaces unobserved entries with zeros and then computes truncated SVD \cite{Cremonesi2010}. We also adapt this model to the cold start settings, as described in \sref{subsec:coldstart}. In addition to that, we implement its scaled variant according to \eref{eq:scaling}. Once the best performing scaling is identified, it is used without further adjustments in HybridSVD as well. We denote the scaled models as \emph{PureSVDs} and \emph{HybridSVDs}.

\emph{Factorization Machines} (FM) \cite{rendle2010factorization} encapsulate side information into interactions via uniform one-hot encoding framework. This is one of the most general and expressive hybrid models. We use implementation by Turi Create\footnote{https://github.com/apple/turicreate} adapted for binary data. It implements a binary prediction objective based on a sigmoid function and includes negative sampling.
The optimization task is performed by stochastic gradient descent (SGD) \cite{bottou2012stochastic} with the learning rate scheduling via ADAGRAD \cite{duchi2011adaptive}.
We also perform \emph{standard MF} by disabling side feature handling in FM. We denote this approach as SGD by the name of the optimization algorithm. It is used to verify whether FM actually benefits from utilizing side features.

\emph{Local Collective Embeddings} model (LCE) \cite{saveski2014lce} is another generic approach that combines the ideas of coupled or collective matrix factorization \cite{singh2008collectivemf, acar2011all} with the Graph Laplacian-based regularization to additionally impose locality constraints. This enforces entities that are close to each other in terms of side features to remain close to each other in the latent feature space as well. We use the variant of LCE where only side features are used to form the Laplacian. We adapted an open-source implementation available online\footnote{https://github.com/abhishekkrthakur/LCE}.

We additionally introduce a heuristic \emph{similarity-based hybrid approach} (SIM), inspired by \eref{eq:folding-in-std}, where the orthogonal projector $\vec{\widetilde V}\vec{\widetilde V}^\top$ can be viewed as item similarity in the latent space. The SIM model simply replaces it with side similarity. In the standard case, it predicts item scores as $\vec{r}=\vec{Sp}$. Likewise, viewing \eref{eq:coldstart} in the same sense gives $\vec{r}=\vec{Rs}$ for user scores prediction in the cold start, where $\vec{s}$ denotes the similarity of a cold-start item to known items.

Finally, we include two non-personalized baselines that recommend either the \emph{most popular} (MP) or \emph{random} (RND) entities.

\subsection{Hyperparameters tuning}
We test all factorization models on a wide range of rank values (i.e., a number of latent features) up to 3000. We use the MRR@10 score for selecting the optimal configuration.

The HybridSVD model is evaluated for 5 different values of $\alpha$ from $\left\{0.1, 0.3, 0.5, 0.7, 0.9\right\}$. Similarly to the PureSVD case and unlike other competing MF methods, once the model is computed for some rank $k_{max}$ with a fixed value of $\alpha$, \emph{we immediately get any model with a lower rank value $k < k_{max}$ by a simple rank truncation} without any additional optimization steps. This significantly simplifies the hyper-parameter tuning procedure as it eliminates the need for expensive model recomputation during the grid search.

In the case of PureSVD we also perform experiments with different values of the scaling parameter $d$ (see \sref{subsec:scaling}) set to 0.2, 0.4, and 0.6. The best performing value is then used for an additional set of experiments with HybridSVD.

Tuning of the FM model includes only the regularization coefficients for the bias terms and for interaction terms. The number of negative samples is fixed to the default value of 4. The number of epochs is set to 25. The initial rank value is set to 100 on Yahoo dataset and 40 on others. Once an optimal configuration for the fixed rank is found, we perform rank optimization.
The hyper-parameter space of the FM model quickly becomes infeasible with the increased granularity of a parameter grid as we do not have the luxury of a simplified rank optimization available in the case of HybridSVD.
In order to deal with this issue, we employ a \emph{Random Search} strategy
\cite{bergstra2012randomsearch} and limit the number of possible hyper-parameters combinations to 60 (except Yahoo!Music dataset, where we set it to 30 due to very long computation time).

Tuning LCE model is similar to FM. In all experiments, we run the algorithm for 75 epochs and use 10 nearest neighbors for generating Graph Laplacian. The initial rank values used for tuning are the same as for FM. We first perform grid search with the values of regularization in the range from 1 to 30 and with the fixed values of the LCE model's coefficients $\alpha$ and $\beta$ set to 0.1 and 0.05 respectively. After this step we tune $\alpha$ and $\beta$ in a range of values from $[0, 1]$ interval (excluding $\alpha=0$). Once the optimal configuration is determined, we finally perform rank values tuning.

\begin{figure*}[ht]
\centering
\includegraphics[width=\textwidth]{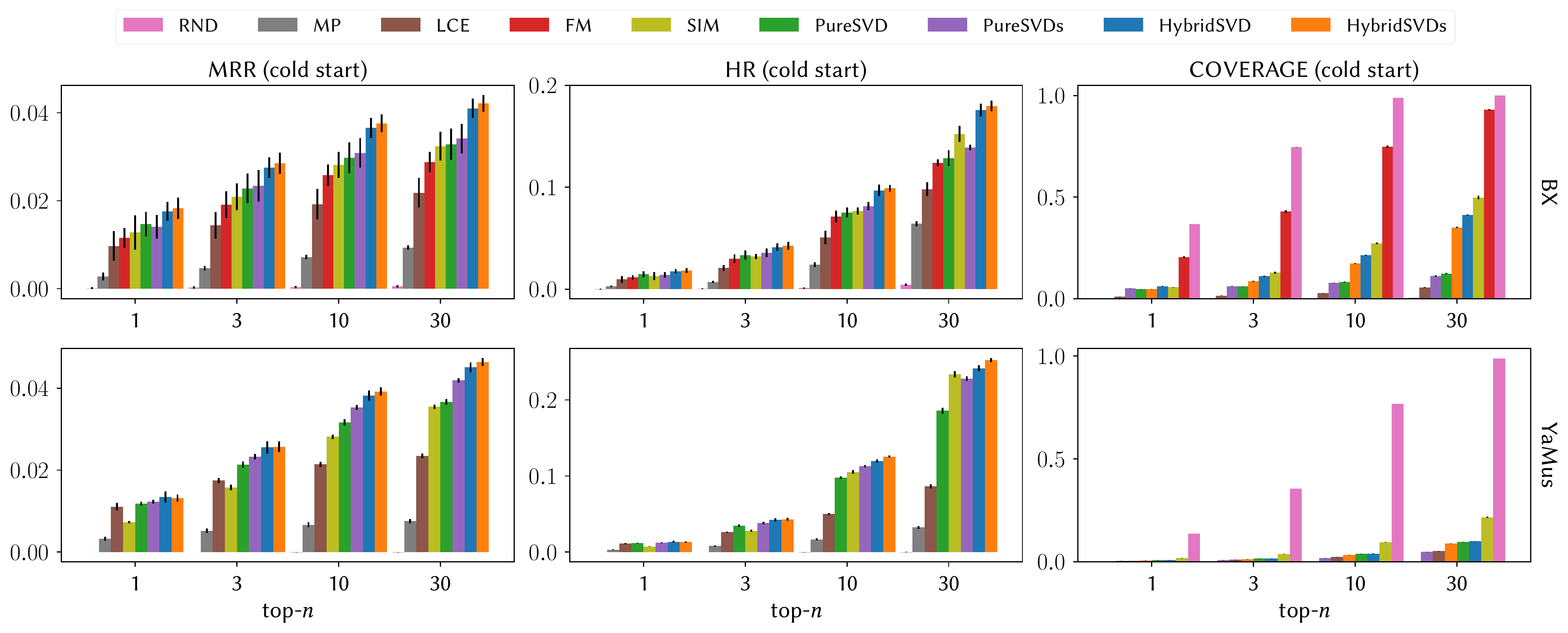}
\caption{Cold start scenario for the BookCrossing (top row) and Yahoo! Music (bottom row) datasets. In contrast to the standard case, the use of both scaling and side information improves predictions quality. The HybridSVD approach significantly outperforms all other methods. Surprisingly, even the cold start variant of PureSVD outperforms some hybrid models.}
\label{fig:cls_mrr}
\end{figure*}

\section{Results and discussion}\label{sec:results}
Results for the standard and the cold start scenarios are depicted in Figures \ref{fig:std_mrr} and \ref{fig:cls_mrr} respectively. Confidence intervals are reported as black vertical bars. For the sake of visual clarity, we do not show all the results and only report the most descriptive parts of it. The complete set of figures, as well as the set of optimal hyper-parameters for each model, can be found in our Github repository\footnote{See Jupyter notebooks named \emph{View\_all\_results} and \emph{View\_optimal\_parameters}.}.

\subsection{Standard scenario}
Let us consider at first the \emph{non-scaled} versions of the algorithms. We observe that, in the standard scenario, there is no absolute winner among hybrid recommenders. Pure collaborative models provide comparable results in 5 out of 6 datasets.

Nevertheless, HybridSVD significantly outperforms the other (non-scaled) algorithms on the YaMus dataset. By a closer inspection of the baselines, we note that this is the only dataset where the SIM model gets remarkably higher relevance scores than the popularity-based model. Probably, it can be explained by a strong attachment of users to their favorite artists or genres. Giving more weight to side features captures this effect and improves predictions, while CF models suffer from popularity biases.
After proper scaling (i.e., debiasing) the effect vanishes, which is indicated by a comparable performance of both PureSVDs and HybridSVDs. Surprisingly, the \emph{FM model was unable to converge to any reasonable solution} on this dataset (we have repeated the experiment 3 times on 3 different machines and obtained the same unsatisfactory result).

Generally, we observed a tight competition between FM, LCE, and HybridSVD. Interesting to note that in some cases non-SVD-based hybrid models were slightly underperforming non-hybrid PureSVD (see ML1M, ML10M results for FM and ML1M, BX results for LCE). In contrast, HybridSVD exhibited more reliable behavior. It can be explained by better control over the weights assigned to side information during the fine-tuning phase, effectively lowering the contribution of side features if it does not aid the learning process. Generally, HybridSVD was performing better than its competitors on 3 datasets -- ML1M, YaMus, and BX; LCE was better than others on 2 datasets -- ML10M and AMZe; FM was better on AMZvg.

A significant boost to recommendations quality of the SVD-based models is provided by the scaling trick (\sref{subsec:scaling}). On almost all datasets the top scores were achieved by the scaled versions of either PureSVD or HybridSVD. The only exception was BX dataset, where the non-scaled version of HybridSVD outperformed the scaled version of PureSVD. The result, however, was not statistically significant and the top-performing model was still the scaled version of HybridSVD. Our results suggest that, in the standard scenario, debiasing data is often more important than adding side information. The effect is especially pronounced in the AMZe and YaMus cases. This also resonates with conclusions in \cite{pilaszy2009recommending}. As the authors argue, ``\emph{even a few ratings are more valuable than metadata}''.

\subsection{Cold start scenario}
In contrast to the standard case, in the cold start scenario, HybridSVD and its scaled version clearly demonstrate superior quality. The scaled version outperforms all other models on 5 out of 6 datasets except AMZvg, where all SVD-based models achieve similar scores (which are still the highest among all other competitors). Even the non-scaled version of HybridSVD significantly outperforms both LCE and FM models on almost all datasets excluding AMZe, where FM performs better than others.

Moreover, unlike the standard scenario, HybridSVD also demonstrates a better quality in comparison with the scaled version of PureSVD on half of the datasets (YaMus, BX, ML1M) and performs comparably well on ML10M and AMZvg. We note that this result could be anticipated by inspecting the baseline models similarly to the standard case: the scores for both scaled and non-scaled variants of HybridSVD are higher than that of PureSVD on the datasets where the SIM model significantly outperforms MP.

Surprisingly and in contrast to HybridSVD, both FM and LCE models significantly underperform even the heuristic-based SIM model on 4 datasets excluding ML10M and AMZe. This does not happen even with the cold start adaptation of the PureSVD approach, which is not a hybrid model in the first place.
The LCE model also has the worst coverage. The SVD-based models, in turn, typically provide more diverse recommendations without sacrificing the quality of recommendations. In general, however, the diversity of recommendations is lower than in the standard case.

We note that the performance of the HybridSVD approach is consistent, mostly favoring the higher values of $\alpha$ (except for the AMZvg case). Unsurprisingly, in the cold start regime, \emph{side features are as important as scaling}, and HybridSVD makes the best use of both of them.
Overall, the proposed HybridSVD approach provides a flexible tool to control the contribution of side features into the model's predictions and to adjust recommendations based on the meaningfulness of side information.

\section{Related work}\label{sec:related}
Many hybrid recommender systems have been designed and explored to date, and a great amount of work has been devoted to incorporating side information into matrix factorization algorithms in particular. We group various hybridization approaches into several categories, depending on a particular choice of data preprocessing steps and optimization techniques.

A broad class of hybrid factorization methods maps real attributes and properties to latent features with the help of some \emph{linear transformation}. A simple way to achieve this is to describe an entity as the weighted sum of the latent vectors of its characteristics \cite{stern2009matchbox}.
In the majority of models of this class, the feature mapping is learned as a part of an optimization process \cite{pilaszy2009recommending,chen2012combining,roy2016latent}. The FM model described earlier also belongs to this group. Some authors also proposed to learn the mapping as a post-processing step using either entire data \cite{gantner2010learning} or only its representative fraction \cite{cohen2017expediting}.

Alternatively, in the \emph{aggregation} approach, feature-based relations are imposed on interaction data and are used for learning aggregated representations. Among notable models of this class are Sparse Linear Methods with Side Information (SSLIM) \cite{SLIM2012}. In some cases, a simpler approach based on the \emph{augmentation} technique can also help to account for additional sources of information. With this approach, features are represented as new dummy (or virtual) entities that extend original interaction data \cite{nabizadeh2016predicting,akhmatnurov2015boolean}.
Another wide class of methods uses \emph{regularization-based} techniques to enforce the proximity of entities in the latent feature space based on their actual features. Some of these models are based on \emph{probabilistic frameworks} \cite{gunawardana2009unified,porteous2010bayesian},
others directly extend standard MF objective with additional regularization terms \cite{nguyen2013content,chen2017leveraging}.

One of the most straightforward and well-studied regularization-based approaches is \emph{collective MF} \cite{singh2008collectivemf}. In its simplest variant, sometimes also called \emph{coupled MF} \cite{acar2011all}, parametrization of side features is constrained only by squared difference terms similarly to the main optimization objective of standard MF \cite{fang2011cofactor}.
There are also several variations of the coupled factorization technique, where regularization is driven by side information-based similarity between users or items \cite{shi2010context,barjasteh2015cold} rather than by side features themselves. As we noted in \sref{subsec:baselines}, the authors of the LCE model extend this approach further with additional Graph Laplacian-based regularization, which is, in turn, related to the kernelized MF \cite{pal2018kernelizedcf,zhou2012kernelizedmf}.

All these methods are based on a general problem formulation, which provides a high level of flexibility for solving hybrid recommendation problems. On the other hand, it sacrifices many benefits of the SVD-based approach, such as global convergence guarantees, direct folding-in computation, and quick rank value tuning achieved by simple truncation of factor matrices.
Considering these advantages, the SVD-based approach has received surprisingly low attention from the hybrid systems perspective. It was shown to be a convenient \emph{intermediate tool} for factorizing combined representations of feature matrices and collaborative data
\cite{symeonidis2008content,ariyoshi2010hybrid}. However, to the best of our knowledge, there were no attempts for developing an integrated hybrid SVD-based approach where interactions data and side information would be jointly factorized without violating the computational paradigm of classical SVD.

Worth mentioning here that SVD generalizations have been explored in other disciplines including studies of genome \cite{alter2003genesvd}, neuroimaging, signal processing, environmental data analysis (see \cite{allen2014gmd} and references therein). Moreover, the authors of \cite{allen2014gmd} propose an elegant computational framework for reducing the dimensionality of structured, dense datasets without explicitly involving square roots or Cholesky factors or any of their inverses. Even though it can potentially be adapted for sparse data, it is not designed for quick online inference, which requires computing matrix roots and the corresponding inverses in any case due to \eref{eq:folding-in}.

\section{Conclusions and future directions}\label{sec:conclusion}
We have generalized PureSVD to support side information by virtually augmenting collaborative data with additional feature-based relations. It allows imposing the desired structure on the solution and, in certain cases, improves the quality of recommendations.

The model is especially suitable for the cold start regime. The orthogonality property of singular vectors admits an easy extraction of the latent representation of side features, required in the cold start, for both PureSVD and HybridSVD. The latter, however, consistently generates predictions of a higher quality. As a result, the proposed method outperforms all other competing algorithms, sometimes by a significant margin.

Conversely, in the standard case, a simple scaling trick often allows achieving a superior quality even without side information. A sufficient amount of collaborative information seems to hinder the positive effect of side knowledge and makes the use of it redundant. Nonetheless, HybridSVD generally provides much more reliable predictions across a variety of cases.

We have proposed an efficient computational scheme for both model construction and recommendation generation in online settings, including warm and cold start scenarios. The model can be further improved by replacing standard SVD with its randomized counterpart. This would not only speed up computations but would also enable support of highly distributed environments.

We have identified several directions for further research. The first one is to add both user- and item-based similarity information into the folding-in setting. One possible way is to combine folding-in vectors with the cold start approximation.
Another interesting direction is to relax the strict positive definiteness constraint and also support compact, dense feature representations, which can be achieved with the help of fast symmetric factorization techniques.

\begin{acks}
The work is supported by RFBR, research project No. 18-29-03187.
\end{acks}

\newpage

\bibliographystyle{ACM-Reference-Format}
\bibliography{bibliography}

\end{document}